\documentclass[runningheads]{llncs}
\usepackage[sorting=none, backend=biber]{biblatex}
\usepackage[super]{nth}
\usepackage[toc,page]{appendix}
\usepackage{placeins}
\usepackage{float}
\usepackage{subcaption}
\usepackage[T1]{fontenc}
\usepackage[english]{babel}
\usepackage{graphicx}

\usepackage{algpseudocode}
\usepackage{algorithm}

\usepackage{hyperref}
\usepackage{csquotes}
\usepackage{amsmath}
\usepackage[super]{nth}
\usepackage[dvipsnames]{xcolor}
\usepackage[x11names]{xcolor}

\usepackage{caption} 
\algnewcommand\algorithmicforeach{\textbf{for each}}
\algdef{S}[FOR]{ForEach}[1]{\algorithmicforeach\ #1\ \algorithmicdo}
\algdef{SE}[REPEATN]{RepeatN}{End}[1]{\algorithmicrepeat\ #1 \textbf{times}}{\algorithmicend}

\AtEveryBibitem{
	\clearfield{urlyear}
	\clearfield{urlmonth}
}

\authorrunning{Deproost et al.}
\titlerunning{Voronoi State Partitioning}

\begin{document}
\title{Explainable RL Policies by Distilling to Locally-Specialized Linear Policies with Voronoi State Partitioning}

\author{ Senne Deproost\inst{1, 2}\orcidID{0009-0009-4757-0290}\and \\
	 Denis Steckelmacher \inst{1, 2}\orcidID{0000-0003-1521-8494}\and \\
	Ann Now\'{e}\inst{1, 2}\orcidID{0000-0001-6346-4564}}

\institute{Vrije Universiteit Brussel, Pleinlaan 2, 1050 Brussels, Belgium \and
	BP\&M, Flanders Make@VUB, Pleinlaan 2, 1050 Brussels, Belgium}
\maketitle              

\begin{abstract}
	Deep Reinforcement Learning is one of the state-of-the-art methods for producing near-optimal system controllers. However, deep RL algorithms train a deep neural network, that lacks transparency, which poses challenges when the controller has to meet regulations, or foster trust. 
	To alleviate this, one could transfer the learned behaviour into a model that is human-readable by design using knowledge distillation. Often this is done with a single model which mimics the original model on average but could struggle in more dynamic situations. A key challenge is that this simpler model should have the right balance between flexibility and complexity or right balance between balance bias and accuracy.
	We propose a new model-agnostic method to divide the state space into regions where a simplified, human-understandable model can operate in. In this paper, we use Voronoi partitioning to find regions where linear models can achieve similar performance to the original controller. We evaluate our approach on a gridworld environment and a classic control task. We observe that our proposed distillation to locally-specialized linear models produces policies that are explainable and show that the distillation matches or even slightly outperforms the black-box policy they are distilled from.
	

\end{abstract}

\section{Introduction}
With the evergrowing demand for complex automation across domains, the need for advanced controllers is outweighing the available engineers that can program them \cite{authority_etal_2025_ReportLabour}.
Optimal control techniques such as Model Predictive Control can optimize a controller given a global model of the system is provided \cite{rawlings_etal_2017_ModelPredictive}. However, in case no model is available or the system is highly dynamic, Deep Reinforcement Learning (DRL) can learn an optimal controller using repetitive interactions with the system \cite{mnih_etal_2015_HumanlevelControl}. DRL has shown state-on-the-art performance in many applications and is being increasingly integrated in the controller generation methodology \cite{bellemare_etal_2020_AutonomousNavigation, degrave_etal_2022_MagneticControl, li_2023_DeepReinforcement}. 
However, the capabilities and strength of DRL comes at the cost of users not understanding how the controller works. The artificial neural network at the core of the controller is complex and gives little insight into why and how it behaves. This raises distrust among users who should implement, test, deploy and operate the controller. 
To alleviate this lack of understanding, the field of \textit{Explainable Reinforcement Learning} (XRL) introduced several solutions \cite{bekkemoen_2023_ExplainableReinforcement}. Here, explanation generating techniques aim to answer questions such as \textit{what} is the global control strategy, \textit{why} a controller behaves in a certain way, \textit{why not} another behaviour, \textit{when} to expect this outcome etc. Often this involves the creation of a surrogate model that is by design more interpretable than the original network while retaining most of the performance. Then, using a technique called knowledge distillation, the surrogate learns post-hoc how to mimic the original model \cite{ji_zhu_2020_KnowledgeDistillation}.
From the myriad of human-readable classes that exist, those based on visual or written decision boundaries are the most informative for controller design. For example, rule-based decision trees construct a hierarchical set of splitting nodes in the state space with a defined behaviour in the leaf nodes. Summarizing the splits along the dimensional axis needed to reach a leaf can be seen as an explanation when that behaviour is performed. This method has shown to produce good tree models that can mimic network behaviour \cite{coppens_etal_2019_DistillingDeep, engelhardt_etal_2023_SamplebasedRule}.
However, we argue that a direct translation to any explainable surrogate limits the types of users that could interact with it. The product life cycle of a controller involves many stakeholders, each with different technical expertise. The control tester is not required to have the same domain specific language (DSL) as the control implementer. A factory worker only needs to interact via the user interface while the operational technician needs to tweak the controller parameters. If instead we would opt for a common language that is explainable to every user, we would omit the use of nuances specific to the user's DSL. Aside communication, fixing a DSL could limit the performance of the controller if the chosen language is unable to capture complex behaviour.
With this shortcoming in mind, we propose an intermediate step between network and explanation. By partitioning the operational state space of the controller into regions with arbitrary simple behaviour, we could provide an interpretable representation that delivers insight for each type of user. 
For an initial version of this idea, we want to find regions where linear functions can operate in with similar performance to the original DRL network. The parameters, or weights, of these models indicate the importance of each input variable in forming the controller output signal while the bias term is an offset factor to this weighted sum. 

\textbf{Contributions:} In this paper, we propose a knowledge distillation algorithm that splits the operational state space of the controller into Voronoi cells. This post-hoc method uses a trained DRL agent to gather experiences in the form of state-action pairs to learn the linear models associated to these cells. The models are continuously optimized and a periodic update is performed to decide their decision boundaries based on their loss in following the original model.
To validate our method, we chose a continuous space gridworld and a control task. We observe the capability our algorithm to find these linear subpolicy regions while staying close to the performance of the original DRL policy.

\section{Background}

\subsection{Reinforcement Learning}


Reinforcement Learning (RL) is a machine learning paradigm for solving sequential decision problems \cite{sutton_barto_2014_ReinforcementLearning}.These are defined as a Markov Decision Process (MDP) and formulated by tuple $ \left ( S, A, \mathcal{P}^a_{ss'},\mathcal{R}^a_{ss'} \right )$ in a control problem scheme \cite{bellman_1957_MarkovianDecision}. Here, the state space $S$ consists of any state $s_t$ that can be observed at $t$ with $t \in \left[0, t_{max} \right]$ while $A$ denotes the action space with $a_t$ the action performed at time $t$. $\mathcal{P}^a_{ss'}$ is the probability distribution for each state transition $s_t \rightarrow s_{t+1}$ when $a_t$ is performed and $\mathcal{R}^a_{ss'}$ associates a reward signal to each transition. Reward is a value indicating how good or bad the chosen actions are in following a certain control objective. The MDP is subject to the Markovian property $Prob\left \{s_{t+1} = s',   r_{t+1} = r  \mid s_t, a_t \right \}$, where the state transition is independent from past transitions.
A policy $\pi_t$ is a mapping $s_t \rightarrow a_t$ that decides the chosen behaviour at each timestep. The objective of the policy is to maximize the discounted sum of rewards $R(\tau) = \sum\limits_{t} \gamma^t r_{t + 1}$, whereby $\gamma \in \left[0, 1\right[$ as a discount factor. 


\subsection{Black Box Deep Reinforcement Learning}
The first RL algorithms represented their policy using tabular values and were only applicable to discrete state and action spaces. Should the policy operate in continuous space, a discretization of the learned values has to be applied with sufficient resolution. How accurate the policy can represent these values is bound by the discretization error, which reduces with table size \cite{busoniu_etal_2010_ReinforcementLearning}. The higher the resolution, the more accurate these values are learned. However, control policies for complex systems often require a small resolution, causing an explosion of table size and therefore system memory.
A solution is to learn these values using an artificial neural network since they are considered universal function approximators and can generalize well over unforeseen inputs \cite{mnih_etal_2015_HumanlevelControl, hornik_etal_1989_MultilayerFeedforward}. For visual inputs, the state could be transformed into an embedding using convolutions before using it as input to a regular feed forward neural network. 
This Deep Reinforcement Learning (DRL) approach has proven itself quite successful in a myriad of applications, especially to control complex dynamic systems. However, transparency of the policy is lost due to the computational complexity of a neural network. The large amount of parameters describing the operations done at inference are only informative on a global level in relation with the parameters of all other layers. In addition, two sets of different parameterized networks can yield similar behaviour rendering the semantic meaning of one parameter useless. 
Recently, the field of Explainable Reinforcement Learning (XRL) has researchers investigate new methods to represent the policy in a more human-understandable manner \cite{bekkemoen_2023_ExplainableReinforcement}. Introduced methods can describe network behaviour on both local and global level.
The method we present aims for a global description using simple linear models. However, we do note that the local description of a region is only given when performing inference with a nearest neighbor algorithm. For a more globally informed explanation, the region should be described in a meaningful way independent from all other regions.

\subsection{Voronoi quantization}

\begin{figure}[t]
	\centering
	\includegraphics[width=.3\textwidth]{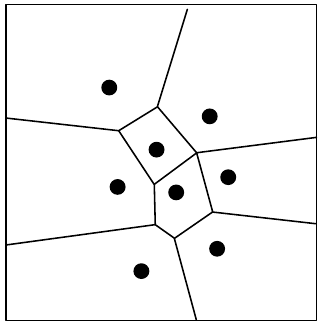}
	\caption{An example of a Voronoi diagram using 8 codeword points.}
	\label{fig:voronoi}
\end{figure}

Vector quantization is a compression technique that groups similar datapoints into regions described by a single representative codeword point $c \in \Re^n$ \cite{gray_1984_VectorQuantization}. A most common one, nearest neighbor vector quantization, defines $c$ to be the point closest to each other point in the region. This quantizer $\psi \rightarrow \Re^n$, the Voronoi quantizer, maps each n-dimensional vector $x \in \Re^n$ onto a finite set of codewords $C = \left\{c_1, c_2, ..., c_m \right\} \subset \Re^n$, both from the set $\Re$ of real numbers. This mapping creates a partitioning of $m$ disjoint Voronoi cells of the vector space following 

\begin{align}
	R_k = \left\{ x \in \Re^n: \psi \left ( x \right ) = c_k \right\}
\end{align}

for any $i = 1, 2, .., m$.

The mapping of a given vector $x$ to a codeword $c_i$ is given by

\begin{align}
	R_k = \left\{ x: \left\| x - c_k\right\| \leq \left\| x - c_j\right\|\right\}, \forall k \neq  j
\end{align}

which implies that $c_i$ is the nearest neighboring codeword of $x$.

The main benefit of this quantization is the simple representation of the regions which is only the list of codewords $C$. The mapping from an arbitrary point to a region $R_k$ can efficiently be done with a nearest neighbor search over all codewords. Often this is done using a kd-tree \cite{maneewongvatana_mount_2001_EfficiencyNearest} which has an average time complexity of $O(\log n)$ to lookup the region-defining codeword.
The intuition behind Voronoi quantization is that a region $R_k$ contains all states that are similar to codeword $c_k$ according to a definition of distance. 
In a later section, we will motivate that the capability of a subpolicy model in following the original global model dictates the selection of these codewords.

\section{Related work}

There are two approaches commonly used in XRL: those that use an inherently interpretable formalism and those that post-hoc imitate a pre-trained policy using state-action pairs. Our approach belongs to the second category. This latter technique, knowledge distillation \cite{hinton_etal_2015_DistillingKnowledge}, has been used to transfer the behaviour from a complex model to a simpler one. Using Soft Decision Trees, decision nodes describe the splits needed to come to a behaviour defined in the leaf nodes \cite{coppens_etal_2019_DistillingDeep}. 
The usage of linear models as an explanation was popularized by Ribeiro et al. using their algorithm LIME \cite{ribeiro_etal_2016_WhyShould}. With it, they can generate linear functions for classification tasks to explain the decision of a model at a local level. In Reinforcement Learning, LIME has been used to explain the behaviour \cite{lu_etal_2025_ExplainableAI} as well as the used reward function \cite{russell_jr_2019_ExplainingReward}.

Our work extends upon earlier work done by Lee et al. \cite{lee_lau_2004_AdaptiveState}. Their online TD-AVQ algorithm was an attempt to use TD-learning in continuous environments. Instead of improving TD-learning, we mainly want to use this idea to find linear models instead of discrete actions, generalizing the behaviour explanation over a wider region compared to a single discrete action.  
 The partitioning algorithm, based on Voronoi cells, aggregates states based on the amount of reward that could be collected. When a certain threshold is exceeded, a new cell is created with the codeword is introduced and the corresponding action is associated with that region. A minimum distance threshold ensured that newly formed regions where not too small, determining the resolution of the partitioning. Additionally, they included a recurring step where a codeword would be removed if neighboring cells had similar action values and therefore can be merged.
Their algorithm has been shown good performance in learning a policy for a navigation gridworld as well as a simple mechanical gripper set up.

\setlength{\tabcolsep}{12pt}
\begin{table}
	\centering
	\begin{tabular}{|c|c|c|}
		\hline
		& TD-AVQ             & Ours            \\
		\hline
		Distillation               & No                 & Yes             \\
		Focus on explainability    & No                 & Yes             \\
		Using modern deep learning & No                 & Yes             \\
		Splitting criteria         & Accumulated reward & Prediction loss \\
		\hline
	\end{tabular}
	\caption{The key differences between TD-AVQ and our method. Whereas TD-AVQ is an improvement on traditional TD-learning, we want to distill interpretable linear policies.}
	\label{tab:comp}
\end{table}

\section{Partitioning the state space}
The main idea behind our approach is to find regions in state space $S$ where a model $\tilde{\pi}$ of arbitrary complexity can produce behaviour that is close to the original policy $\pi$. In this paper, we limit ourselves to linear models but note that this method is model-agnostic.

In algorithm \ref{alg:voronoi-partitioning} we describe the main loop that in each iteration both trains the subpolicies and manages the partitions of the state space. The loop runs for $n_\texttt{epochs}$ and performs operations on the partitions at different iteration frequencies.
Every $n_\texttt{split}$ iterations, the algorithm finds regions that are eligible to be split due to their performance being insufficient on a given trajectory. Every $n_\texttt{merge}$ iterations, each region is compared to its neighbors to check whether or not their learned subpolicy is similar and if one of them can be removed. The last $n_\texttt{freeze}$ iterations do not alter the partitioning but instead only optimize the subpolicies. 

 The list $\tilde{\pi}$ is initially comprised of only one subpolicy $\tilde{\pi}_0$ with its parameters set arbitrarily. The list of codewords $C$ has only $c_0$, which is the first state $s_0$ that will be observed in the first iteration. To map an observed state $s_t$ to a subpolicy $\tilde{\pi}_i$, a kd-tree \cite{maneewongvatana_mount_2001_EfficiencyNearest} is constructed using the list of all codewords $C$. With the state, inference $\texttt{kd-tree}(C, s_t)$ is performed to find the single nearest neighbor codeword based on Manhattan distance. The returned index $i$ corresponds with the index of the policy in the subpolicies list $\tilde{\pi}$.

\begin{algorithm}
	\caption{Voronoi State Partitioning}
	\begin{algorithmic}[1]
		
		\Require Trained policy $\pi$, environment \texttt{env}, empty buffer $B$
		\State Initialize list of subpolicies $\tilde{\pi}$ with policy $\tilde{\pi}_0$ and $s_0$ as codeword $c_0 \in C$
		\For{$n = 0$ to $n_\texttt{epochs}$}
		
		\For{$t = 0$ to $t_{max}$}									\Comment{Collect experiences}
		\State $a_t \gets \pi(s_t)$
		\State Perform $a_t$ in \texttt{env}, observe $s_{t+1}$
		\State $i = \texttt{kd-tree}(C, s_t)$						\Comment{Find nearest codeword}
		\State Add $s_t$ and $a_t$ to buffers $\tilde{\pi}_i$
		\State Add $s_t$to $B$
		\If{$\operatorname{terminal}(s_t)$}
		\State End episode
		\EndIf 
		\EndFor
		
		\ForEach{$\tilde{\pi}_i \in \tilde{\pi}$}					\Comment{Train subpolicies with their buffers}
		\State $\texttt{train} (\tilde{\pi}_i)$
		\EndFor
		
		\If{$n < n_\texttt{freeze}$}									
		
		\ForEach{$n_\texttt{split}$ iteration}						\Comment{Split regions}
		
		\State $\texttt{split\_regions}(C, \tilde{\pi})$
		
		\EndFor
		
		\ForEach{$n_\texttt{merge}$ iteration}							\Comment{Merge regions}
		
		\State $\texttt{merge\_regions}(C, \tilde{\pi})$

		\EndFor
		
		\EndIf
		
		\State $\texttt{reset}(B)$
		
		\EndFor
		
	\end{algorithmic}
	\label{alg:voronoi-partitioning}
\end{algorithm}

\subsection{Learning subpolicies}
Each iteration $n$, a full episode in the environment is performed with RL policy $\pi$ to gather both state transitions $s_t \gets s_{t+1}$ and selected actions $a_t$. Each state $s_t$ is orderly stored in buffer $B$ for 1 iteration to be used as trajectory for the splitting of regions. At each step, a lookup is performed to find the closest code word with the \texttt{kd-tree} and both state and action are added to the buffer of the corresponding subpolicy.
After every episode, each subpolicy is trained using a MSE loss with mini batches of their corresponding buffers. We note that we deliberately want to overfit the models since newly seen data should be captured by new subpolicies if needed. 
The last $n_\texttt{split}$ iterations are used to only train the subpolicies and not to alter the partitioning. This is done to avoid lower performance of newly introduced subpolicies right before the final iteration is reached.

\subsection{Splitting regions}
Splitting one region $c_i$ into two is based on the performance of subpolicy $\tilde{\pi}_i$ when mimicking $\pi$ in that region (Alg. \ref{alg:split}). 
Every $n_\texttt{split}$ episodes, the gathered episode trajectory of $\pi$ is traversed a second time using actions from $\tilde{\pi}$. 
For each encountered state $s_t$ that is part of a new region $c_i$, the corresponding subpolicy with index $i = \texttt{kd-tree}(C, s_t)$ is retrieved and a list of regional losses $\texttt{loss}_{\tilde{\pi}_i}$ is initialized empty. 

With every state that is contained within the region, inference with both $\pi(s_t)$ and $\tilde{\pi}_i(s_t)$ is performed. Loss is calculated as the mean squared error (MSE) between action $a^{\pi}_t$ and $a^{\tilde{\pi}_i}_t$ and added to $\texttt{loss}_{\tilde{\pi}_i}$. 

If a state is part of a different region, $\texttt{loss}_{\tilde{\pi}_i}$ is reinitialized to 0. When the mean of gathered regional losses exceeds threshold value \texttt{max\_loss} at $s_t$ and the distance between that state and regional codeword $c_i$ exceeds \texttt{min\_pol\_distance}, a split is performed. A new subpolicy is initialized randomly and added to $\tilde{\pi}$ and $s_t$ becomes a new codeword $c_j$ added to $C$. Since this split effects all bordering regions, the neighbours of the old region are found using \textit{Delaunay Triangulation} \cite{delaunay_1934_SphereVide} and their buffers are reset to avoid experiences that would be under control of the newly formed policy. 

\begin{algorithm}
	\caption{\texttt{split\_regions}}
	\begin{algorithmic}[1]
		\State $i_\texttt{prev}$ = \texttt{kd-tree}$(C, s_0)$
		\State $\texttt{loss}_{\tilde{\pi}_i} = 0$
		\For{$s_t \in B$}
		\State $i$ = \texttt{kd-tree}$(C, s_t)$
		
		\If{$ i \neq i_\texttt{prev}$}
		\State $i_\texttt{prev} = i$
		\State $\texttt{loss}_{\tilde{\pi}_i} = 0$
		\EndIf
		
		\State $\texttt{loss}_{\tilde{\pi}_i} \gets \texttt{loss}_{\tilde{\pi}_i} \cup \operatorname{MSE}(\pi(s_t), \tilde{\pi}_i(s_t))$
		
		\If{$\operatorname{mean}(\texttt{loss}_{\tilde{\pi}_i}) > \texttt{pol\_loss}_\texttt{max}$ and $ \left\| s_t - c_i \right \|  > \texttt{min\_pol\_disance}$}
		\State Add $s_t$ to $C$
		\State Add new $\tilde{\pi}_i$ to $\tilde{\pi}$
		\State $M \gets \operatorname{neighbours}(\tilde{\pi}_i)$
		\State $\texttt{reset\_buffers}(M)$
		\EndIf
		\EndFor
	\end{algorithmic}
	\label{alg:split}
\end{algorithm}

The intuition behind this approach is that a trained subpolicy is only capable of performing behaviour of a certain complexity in a region of the state space. If the policy performs good enough, it can handle these states. If the loss starts increasing, it has difficulties to perform the behaviour at this moment in its training. If however too much loss is accumulated, we could see that moment as the first encountered most difficult region. Adding another subpolicy at that point gives the other subpolicy a better demarcated region of state space where it has proven itself before. The newly introduced subpolicy starts with empty buffers and initial learning conditions.

\subsection{Merging regions}
When performing splits, the newly chosen codewords could be sub-optimal since we could further train the affected subpolicies. If a split occurs in regions where the behaviour of the original policy differs little, neighbouring subpolicies could emerge with similar behaviour and parameters. To avoid this, every $n_\texttt{merge}$ iterations a pairwise comparison between each subpolicy and its neighbors $M$ is performed.
As a measurement of similarity, we use the $L\infty$ norm of both sets of subpolicy parameters. If this norm is above a certain threshold, the subpolicies are different enough to be kept. If this value is below this threshold, the \texttt{merge\_regions} procedure merges the regions (Alg. \ref{alg:merge}). This removes subpolicy $\tilde{\pi}_j$ and associated codeword $c_j$ from the known regions. The buffer of the remaining subpolicy $\tilde{\pi}_i$ is reset together with all neighboring subpolicies in $M$ since their decision boundaries are impacted as well.

\begin{algorithm}
	\caption{\texttt{merge\_regions}}
	\begin{algorithmic}[1]
		\ForEach{$\tilde{\pi}_i \in \tilde{\pi}$}
		\State $M \gets \operatorname{neighbours}(\tilde{\pi}_i)$
		\ForEach{$j \in M$}
		\If{$\left\| \tilde{\pi}_i.parameters_n - \tilde{\pi}_j.parameters_n \right\|_{\infty} < \texttt{min\_param\_distance}$}
		\State Remove $\tilde{\pi}_j$ from $\tilde{\pi}$
		\State Remove $c_j$ from $C$
		\State $\texttt{reset\_buffers}(M)$
		\EndIf
		
		\EndFor
		\EndFor
	\end{algorithmic}
	\label{alg:merge}
\end{algorithm}


\section{Experimental Validation}

\begin{figure}[H]
	\begin{subfigure}{.5\textwidth}
		\centering
		\includegraphics[width=.65\textwidth]{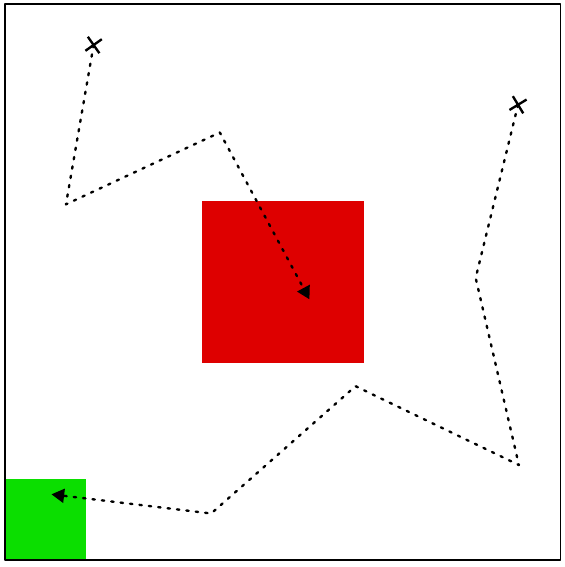}
		\caption{SimpleGoal}
		\label{fig:simplegoal_env}
	\end{subfigure}
	\begin{subfigure}{.5\textwidth}
		\centering
		\includegraphics[width=.94\textwidth]{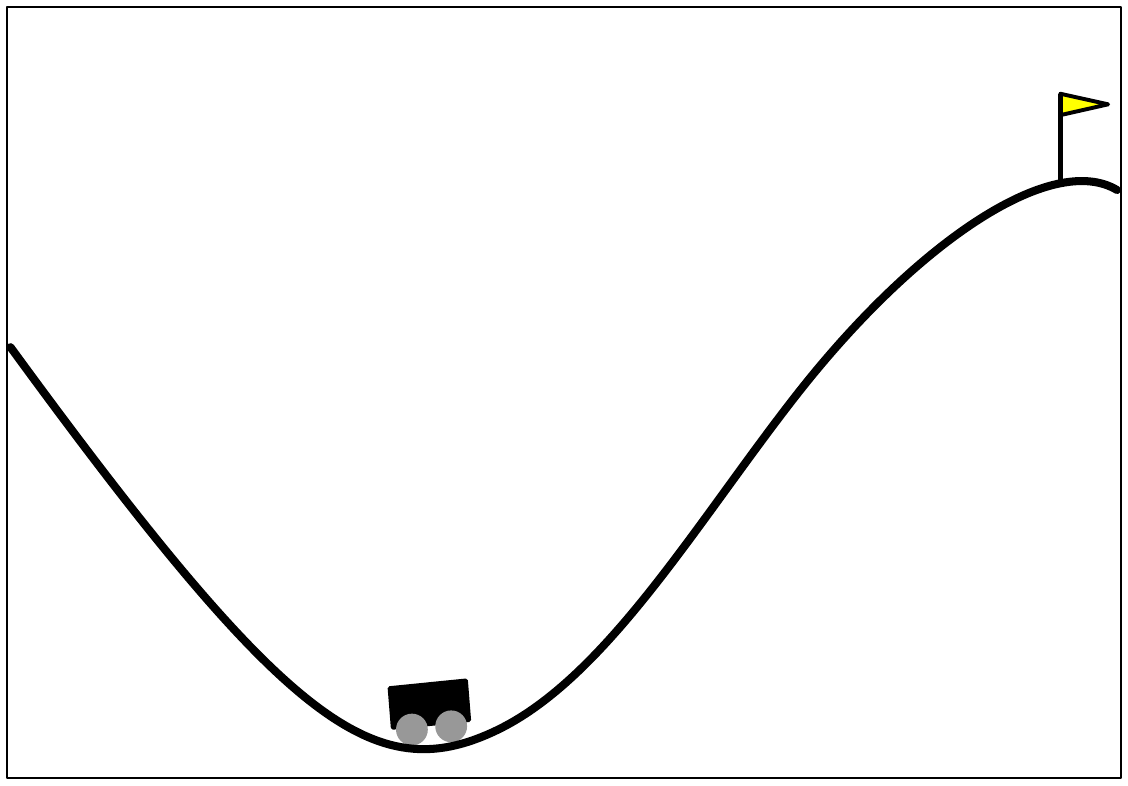}
		\caption{MountainCarContinuous}
		\label{fig:mountaincar_env}
	\end{subfigure}
	\caption{The two environments used to validate our method.}
\end{figure}

To validate our approach, we used both a continuous space gridworld, called \texttt{SimpleGoal}, as a navigation and \texttt{MountainCarContinuous} from the Gymnasium library as a classic control problem \cite{moore_1990_EfficientMemorybased, towers_etal_2024_GymnasiumStandard}. As the original policy, we used a standard version of TD3, a well-known DRL algorithm \cite{fujimoto_etal_2018_AddressingFunction}. For \texttt{SimpleGoal} we trained a TD3 agent for 500.000 steps using the standard parameters while for \texttt{MountainCar} we used a pretrained one from the Stable-Baselines 3 / RL Zoo Hugging Face repository \cite{raffin_2020_RLBaselines3, raffin_etal_2021_Stablebaselines3Reliable}. 

Evaluation is done by analyzing the spread of the gathered episodic returns. To account for the stochasticity  of our distillation  approach we use a DRL policy to obtain 85 distilled policies policies for \texttt{SimpleGoal} (i.e we apply  algorithm \ref{alg:voronoi-partitioning}, 85 times using the same DRL policy) and 40 for \texttt{MountainCarContinuous}. We evaluate each of the distilled policies for 1000 runs using random start states and observe the episodic returns. This yields a spread of 85000 and 40000 individual runs that we use to compare the DRL's 1000 evaluations with. 
Outliers are not shown on the boxplot but are analyzed afterwards.
Since both environments have an action space of size 2, we can visualize the produced policies for an improved understanding how the algorithm performs the partitioning.

Training the DRL policy and subpolicies were done on a M2 MacBook Pro with 16GB of RAM. The used deep learning library is PyTorch v2.3.1 in a Python 3.11.9 environment using Adam as the parameter optimizer \cite{kingma_ba_2015_AdamMethod, paszke_etal_2019_PyTorchImperative}.

\subsection{Navigation task}
\subsubsection{Description}
In \texttt{SimpleGoal} (Fig. \ref{fig:simplegoal_env}) has to navigate in continuous space towards a goal region while avoiding a pitfall in the middle. The task is performed in a bounded space $1.0 \times 1.0$ with the goal being located at $x < 0.1, y < 0.1$ and the pitfall at $0.4 < x < 0.6, 0.4 < y < 0.6$. The observation space is the current $x$ and $y$ coordinate of the agent. The action, with space bound by $[-1, 1]$, is the change in $x$ and $y$ for the next step and is calculated by $dx = 0.1 a_0$ and $dy = 0.1 a_1$. Each timestep, a reward of $r_t = 10 * (\texttt{old\_distance} - \texttt{new\_distance})$ is returned based on the progress the agent makes in approaching the goal. An additional reward of 10 is given if the goal is reached within the truncation time of 50 steps. When the agent enters the pitfall, a reward of $-10$ is given and the episode is terminated.

\begin{figure}[t]
		\centering
		\includegraphics[width=0.65\textwidth]{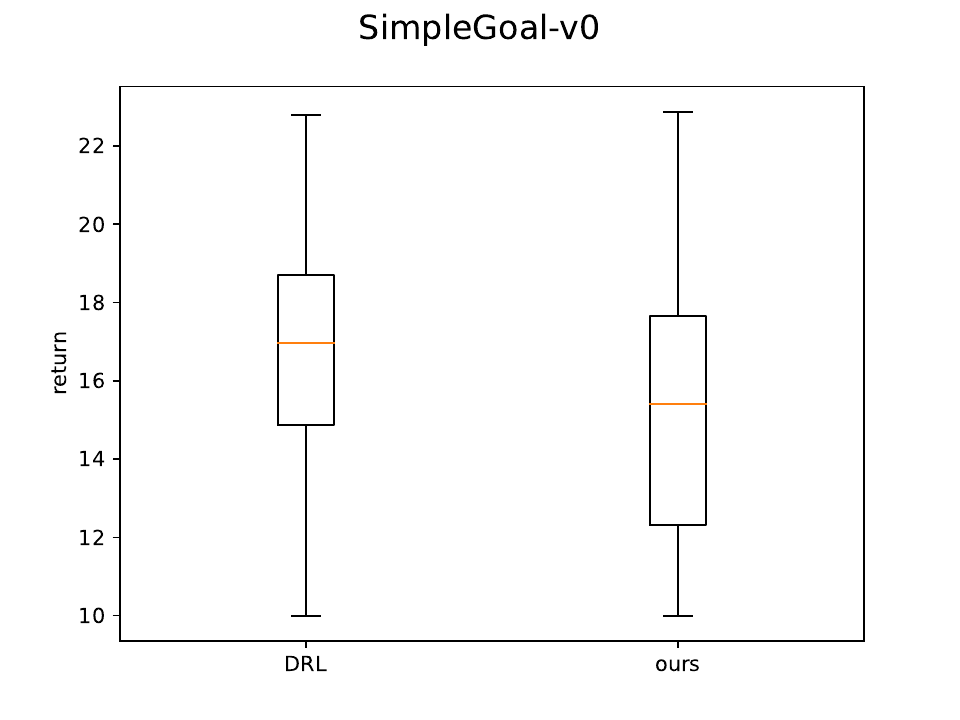}
		\label{fig:simplegoal_boxplot_all}
	\caption{Spread of performance in terms of achieved return (higher is better). The DRL agent is evaluated on 1000 evaluations while 85 instances of our method were evaluated on the same 1000 episodes for a total of 85000 runs. We observe similar spread and performance in both settings.}
	\label{fig:simplegoal_boxplots}
\end{figure}

\subsubsection{Performance} The performance of our algorithm for $40 \times 1000$ runs is reported in figure \ref{fig:simplegoal_boxplots} (right part) and compared to the 1000 runs of the DRL agent (left part).
The DRL agent has returns in a range of $[-10.0, 22.667]$ with mean return 16.599 standard deviation 3.117. 24 outliers have been observed, with a collective mean of 8.369 and standard deviation 5.539. They are spread over range $[-10.0, 10.364]$.
For performance on all runs, we observe an average return of 12.657 with standard deviation 8.436.
 with values ranging $[-11.790, 22.798]$. The outliers, in total 11067, have a mean of -8.042 and a standard deviation of 1.830. They spread over a range of $[-11.790, 1.192]$.\\

If we summarize the overall performance of the method, we see it can learn the \texttt{SimpleGoal} environment with a slight decrease in performance. The median over all training sessions is lower but its spread is similar to the DRL agent. The observed outliers over all runs, often in the negative range, indicate that the number of linear policies in decisive regions such as around the pitfall should be increased or the behaviour of existing subpolicies should be more complex to avoid negative reward being encountered.   

\subsubsection{Visualization}
We visualize the behaviour of each policy using a quiver plot where each arrow originates from a state and points into the direction of movement indicated by the policy (Fig. \ref{fig:simplegoal_polviz}). The linear functions that makes up this partitioning can be found in appendix \ref{appendix:simplegoal_functions}.

We notice that, globally, the plot of the subpolicies tends to follow the original DRL one. The 13 partitions of the state space are distributed seemingly uniformly with key regions at the goal state and around the pitfall. The borders between cells indicate a sudden difference in behaviour, something we could observe in the DRL one as well. The general trend is a movement towards the goal with behaviour to go around the pitfall region. However, we notice that in several important states around the pitfall the agent fails to avoid it.

\begin{figure}[t]
	\begin{subfigure}{.5\textwidth}
		\centering
		\includegraphics[width=0.9\textwidth]{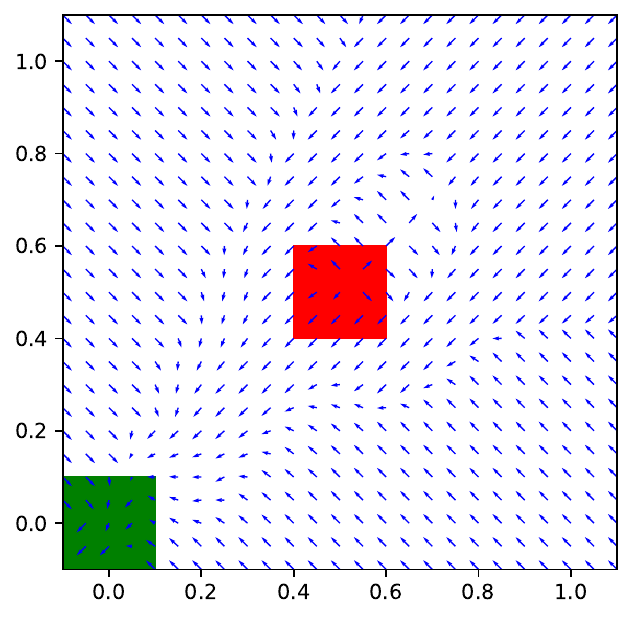}
		\caption{TD3}
		\label{fig:simplegoal_polviz_drl}
	\end{subfigure}
	\begin{subfigure}{.5\textwidth}
		\centering
		\includegraphics[width=0.9\textwidth]{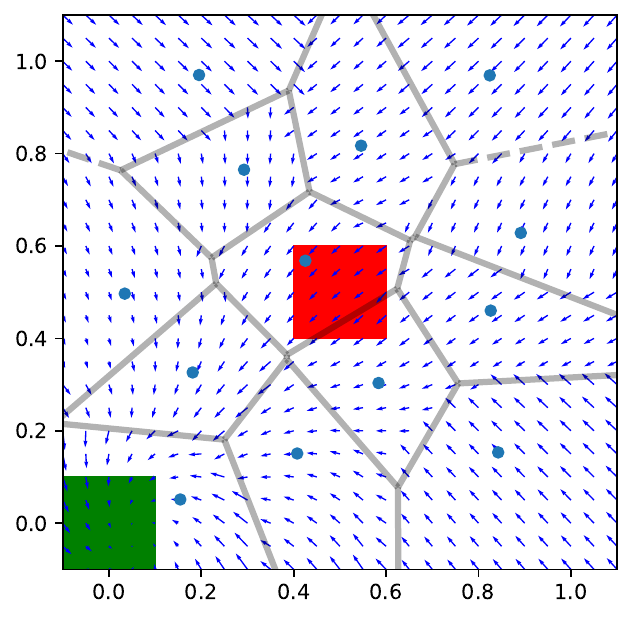}
		\caption{Distilled 13 locally-linear policies.}
		\label{fig:simplegoal_polviz_ipr}
	\end{subfigure}
	\caption{On SimpleGoal: original black-box policy learned with TD3, and the result of its distillation to explainable locally-linear policies.}
	\label{fig:simplegoal_polviz}
\end{figure}


\subsection{Control task}
\subsubsection{Description} \texttt{MountainCarContinuous} is a classic control problem where a toy car has to climb out of a sinusoidal valley towards the top of a hill. The observation space is the position of the car $x \in [-1.2, 0.6]$ along the x-axis and the velocity $v\in [-0.07, 0.07]$. The action space is continuous and its value, applied force $F$, is bound by $F \in [-1, 1]$. The goal state is located at $x = 0.45$ on top of the hill. The truncation length of an episode is 1000 steps.
For each step, a reward of $-0.1*F^2$ is returned, which penalizes the agent applying performing large forces on the car. When the goal is reached, a reward of 100 is given and the episode is terminated with success.

\begin{figure}[t]
		\centering
		\includegraphics[width=0.65\textwidth]{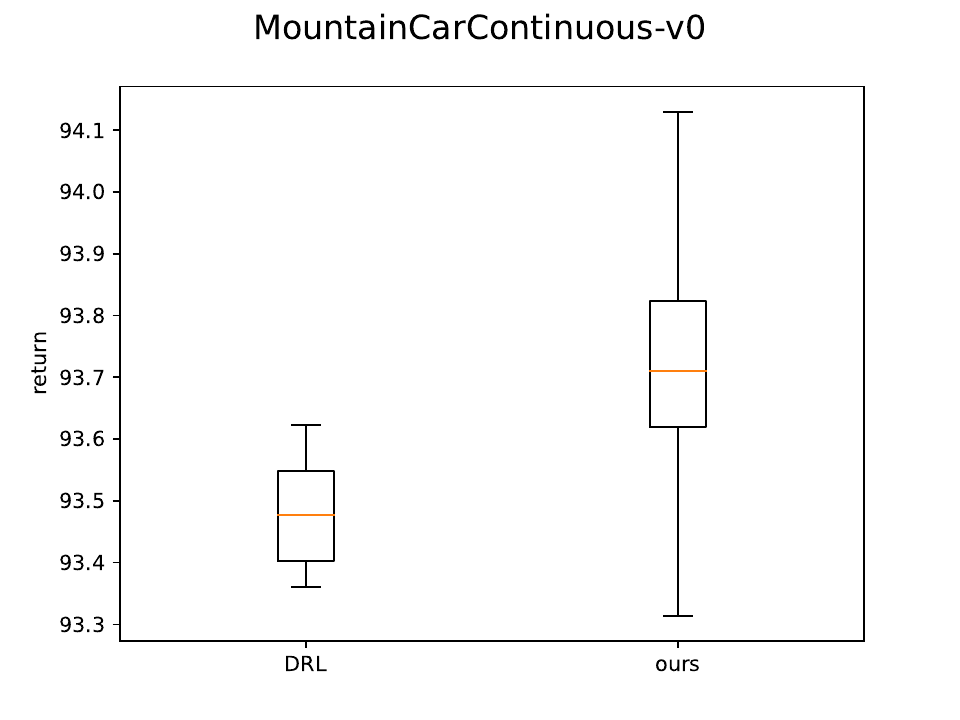}
		\label{fig:mountaincar_boxplot_all}
	\caption{Spread of performance in terms of achieved return (higher is better). The DRL agent is evaluated on 1000 evaluations while 40 instances of our method were evaluated on the same 1000 episodes for a total of 40000 runs. We observe a larger spread in our method and a higher median performance compared to the DRL agent.}
		\label{fig:mountaincar_boxplots}
\end{figure}

\subsubsection{Performance}
The $85 \times 1000$ runs of our algorithm are reported in figure \ref{fig:mountaincar_boxplots} (right part) and compared to the 1000 runs of the DRL agent (left part).
TD3 performs over all episodes with a mean return of 93.481 and a standard deviation of 0.075 within a range of $[93.360, 93.622]$. No outliers were observed.
For all runs with our method, we observe a mean of 93.534 with standard deviation of 1.345 and a range of $[71.050, 94.680]$. 3130 outliers are spread with $73.19\%$ of values below the spread lowerbound of 93.313 and $26.81\%$ above upperbound of 94.130. Respective means are 90.116 and 94.221 with standard deviation 4.338 and 0.110. \\

Over all runs, we observe a higher median compared to DRL. When looking at the mean per run of the algorithm, we observe that the partitioning consistently produces policies with higher returns. We do note that the difference is quite small relative to the return gained by DRL.
The outliers indicate training where the algorithm did not cover the space well enough but still manages to fulfill the task (a positive return always indicates success).

\subsubsection{Visualization}
\begin{figure}[t]
	\begin{subfigure}{.5\textwidth}
		\centering
		\includegraphics[width=.9\textwidth]{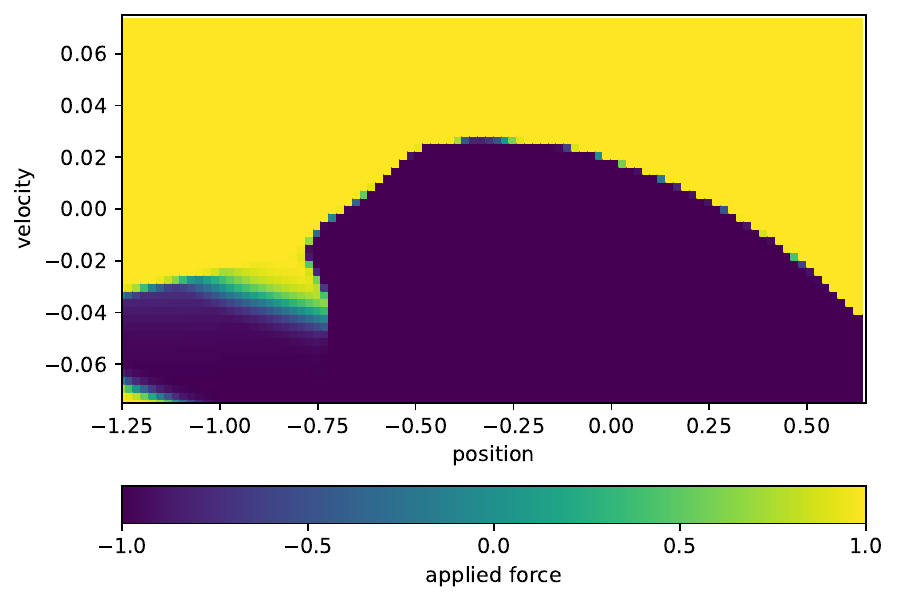}
		\caption{TD3}
		\label{fig:mountaincar_polviz_drl}
	\end{subfigure}
	\begin{subfigure}{.5\textwidth}
		\centering
		\includegraphics[width=.9\textwidth]{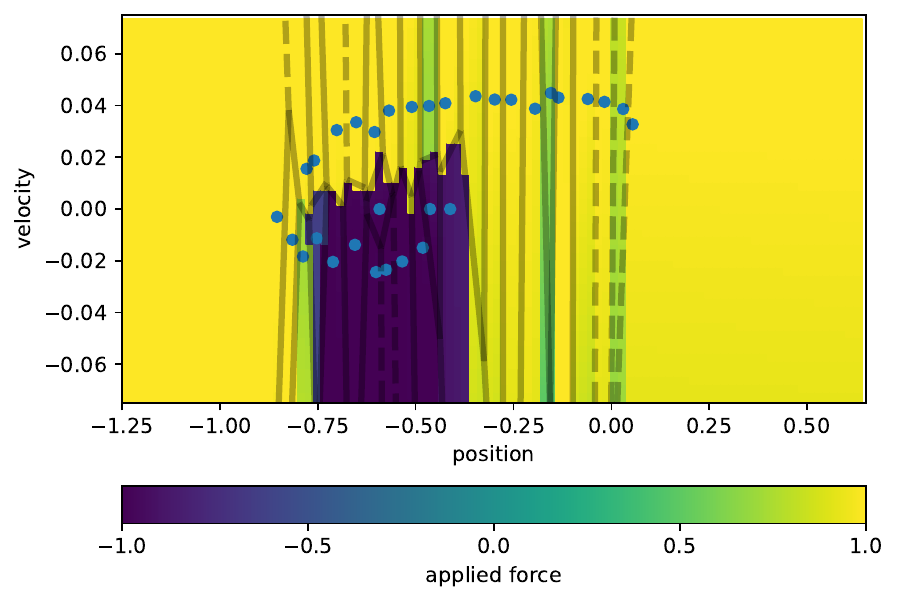}
		\caption{Ours}
		\label{fig:mountaincar_polviz_ipr}
	\end{subfigure}
	\caption{On MountainCar: original black-box policy learned with TD3, and the result of its distillation to explainable locally-linear policies.}
	\label{fig:mountaincar_polviz}
\end{figure}

Both the DRL policy and best subpolicies are visualized using a hotmap over their state space (Fig. \ref{fig:mountaincar_polviz}). The x-axis indicates the position of the car while velocity is given on the y axis. At each point, the color indicates the amount of force applied on the car. 
The 32 regions of the partitioned space have their defining codeword in a spiral-like shape that closely follows the hill of the DRL policy at higher velocities. The region with negative force (colored blue to purple) is much smaller with the linear functions and has a slightly higher value (Appendix \ref{appendix:mountaincar_functions}). We also notice that the decision boundary between negative and positive force is more rigid compared to DRL.

\section{Discussion}
In this paper, we presented a new representation of a Deep Reinforcement Learning using Voronoi State Partitioning. By searching regions where a simple linear model can perform well enough to closely follow a DRL policy, we are able to have global insight in the behaviour of the policy. We validated the approach on both a navigation task and a control task and observed promising, even improved, results compared to the original policy.

Both validation environments have an observation space dimensionality of 2. However, for partitioning a state space with high dimensionality, the kd-tree lookup becomes exponentially less efficient with the number of dimensions \cite{maneewongvatana_mount_2001_EfficiencyNearest}. This \textit{curse of dimensionality} also affects lookup accuracy since with high dimensions, an observation that is equal on all dimensions but one could be considered similar. The definition of nearest points in space doesn't hold anymore.

The chosen hyperparameters were retrieved ad-hoc. A more rigid search for a better learning setting would yield an improved coverage of subpolicies over the state space.

Interpretability, and eventually explainability, of a DRL policy is our main motivation for designing this algorithm. However, due to the way regions are constructed by codeword points and their position in state space relative to other points, it becomes difficult to interpret the demarcation of a region. The method allows for the interpretation of the policy only if inference is performed on the Voronoi cells using a kd-tree. A future version should instead define regions along the axis of the state space, something rule-based and tree ensembles are capable off.

At last, we emphasize the model-agnostic nature of the algorithm. Every class of model with learnable parameters could be used to stand in for regional behaviour. In a later iteration of this approach, we would investigate more controller like schemes such as PID controllers.

\begin{credits}
	\subsubsection{Acknowledgements} This research received funding from the Flanders Research Foundation via FWO S007723N (CTRLxAI) and FWO G062819N (Explainable Reinforcement Learning). We acknowledge financial support from the Flemish Government (AI Research Program).
	
	\subsubsection{Disclosure of Interests} The authors of this dissemination declare to have no conflict of interest with any other party.
	
	\subsubsection{Reproduction} Code is made available for reproduction at \url{https://gitlab.ai.vub.ac.be/sdeproost/IPR.git} under MIT license.
	
\end{credits}
\vfill
\pagebreak
\printbibliography

\vfill
\pagebreak

\appendix

\section{Hyperparameters}
\setlength{\tabcolsep}{12pt}

\begin{table}
	\centering
	\begin{tabular}{|c|c|c|}
		\hline
		Hyperparameter                                 & SimpleGoal-v0   			& MountainCarContinuous-v0\\			
		\hline
		$n_\texttt{epochs}$                            & 5000						& 2000\\
		$n_\texttt{split}$                             & 20							& 50\\
		$n_\texttt{merge}$                             & 100						& 100\\
		$n_\texttt{freeze}$                            & 1000						& 400\\
		$n_\texttt{reset}$                             & 500						& 500\\
		$\texttt{min\_param\_distance}$                & 0.5						& 0.3\\
		$\texttt{min\_pol\_distance}$                  & 0.3						& 0.04\\
		$\texttt{max\_pol\_loss}$                      & 0.0001						& 0.00001\\
		\texttt{one\_split}                            & False						& False\\
		\hline
	\end{tabular}
	\caption{Used hyperparameters for each environment}
\end{table}
\label{appendix:hparams}

\newpage

\section{Best linear policies}

\subsection{SimpleGoal}



\begin{table}
	\centering
	\begin{tabular}{|c|c|c|}
		\hline
		Codeword                    & $\Delta x$  			  &$\Delta y$     \\						
		\hline
		\texttt{[0.891, 0.628]}		&$-0.148x-0.021y-0.055$&$-0.420x+0.231y-1.095$ \\
		\texttt{[0.826, 0.460]}		&$-0.347x+0.305y-0.212$&$-1.087x-1.370y-0.319$ \\
		\texttt{[0.407, 0.150]}		&$3.175y-1.000$&$-4.127y-0.710$\\	
		\texttt{[0.181, 0.326]}		&$-4.588x+0.045y-0.620$&$3.134x-0.082y-0.978$\\	
		\texttt{[0.425, 0.568]}		&$-1.267x+0.966y-0.056$&$-0.281x-0.967y-0.702$\\
		\texttt{[0.292, 0.765]}		&$-1.657x-0.271y+0.147$&$0.433x-0.602y-0.484$\\
		\texttt{[0.154, 0.051]}		&$-5.328x+5.256y+0.191$&$-2.583x-4.501y-0.461$\\
		\texttt{[0.545, 0.817]}		&$-0.124x-0.035y-0.964$&$0.082x-1.027y+0.226$\\
		\texttt{[0.842, 0.153]}		&$-0.628x+0.406y-0.385$&$-0.536x-0.010y+0.663$\\
		\texttt{[0.034, 0.496]}		&$-0.649x-0.652y-0.076$&$0.706x-0.6946y-0.463$\\
		\texttt{[0.583, 0.303]}		&$-0.290x-0.470y-0.855$&$0.103x-3.580y+0.991$\\
		\texttt{[0.824, 0.969]}		&$0.266x-0.424y-0.684$&$-0.598x-0.256y-0.420$\\
		\texttt{[0.195, 0.970]}		&$0.327x-0.246y+0.349$&$0.620x-0.077y-0.882$\\
		\hline
	\end{tabular}
	\caption{All codewords and subpolicies, rounded to 4 digits.}
\end{table}
\label{appendix:simplegoal_functions}

\newpage

\subsection{MountainCar}



\begin{table}
	\centering
	\begin{tabular}{|c|c|}
		\hline
		Codeword                    & $F$  		     \\						
		\hline
		\texttt{[-0.592, 0.000]} & \color{BlueViolet} $-0.375x+3.004v-1.205$\\
		\texttt{[-0.463, 0.000]} & \color{BlueViolet} $1.664x+2.371v-0.211$\\
		\texttt{[-0.510, 0.040]} & \color{Goldenrod}  $-1.117x+1.050v+0.340$\\
		\texttt{[-0.568, 0.038]} & \color{LimeGreen}   $-0.707x+0.454v+0.585$\\
		\texttt{[-0.575, -0.024]}& \color{BlueViolet} $0.8357x+0.7049v-0.514$\\
		\texttt{[-0.154, 0.045]} & \color{Goldenrod}  $-0.952x+1.840v+0.470$\\
		\texttt{[-0.655, -0.014]}& \color{BlueViolet} $-0.007x+0.840v-0.992$\\
		\texttt{[-0.424, 0.041]} & \color{Goldenrod}  $-1.270x+1.317v+0.417$\\
		\texttt{[-0.298, 0.042]} & \color{LimeGreen}   $-0.840x+1.083v+0.701$\\
		\texttt{[-0.256, 0.042]} & \color{Goldenrod}  $-0.216x+0.673v+0.916$\\
		\texttt{[-0.135, 0.043]} & \color{Goldenrod}  $-0.97x-0.264v+0.888$\\
		\texttt{[-0.018, 0.041]} & \color{Goldenrod}  $-0.435x+0.738v+0.965$\\
		\texttt{[-0.711, -0.020]}& \color{BlueViolet} $0.867x+0.415v-0.374$\\
		\texttt{[-0.481, -0.015]}& \color{BlueViolet} $1.125x+0.788v-0.440$\\
		\texttt{[-0.854, -0.003]}& \color{Goldenrod}$-1.006x+0.301v+0.150$\\
		\texttt{[-0.060, 0.043]} & \color{Goldenrod}  $-1.115x+1.334v+0.867$\\
		\texttt{[-0.788, -0.018]}& \color{Goldenrod}  $-0.416x-0.778v+0.390$\\
		\texttt{[-0.815, -0.012]}& \color{Goldenrod}  $-0.569x+0.090v+0.535$\\
		\texttt{[-0.778, 0.016]} & \color{Goldenrod}  $-0.774x+1.08v+0.376$\\
		\texttt{[-0.347, 0.044]} & \color{Goldenrod}  $-1.147x+0.772v+0.521$\\
		\texttt{[0.030, 0.039]}  & \color{LimeGreen}   $1.199x+0.872v+0.734$\\
		\texttt{[-0.752, -0.011]}& \color{Violet}     $0.871x-0.737v+0.018$\\
		\texttt{[-0.604, 0.030]} & \color{Goldenrod}  $-0.281x+0.511v+0.811$\\
		\texttt{[-0.194, 0.038]} & \color{Goldenrod}  $-1.299x+0.470v+0.705$\\
		\texttt{[0.054, 0.033]}  & \color{Goldenrod}  $0.007x+0.701v+0.974$\\
		\texttt{[-0.412, 0.000]} & \color{BlueViolet} $1.0258x+0.073v-0.464$\\
		\texttt{[-0.601, -0.024]}& \color{BlueViolet} $0.7532x+1.528v-0.514$\\
		\texttt{[-0.701, 0.030]} & \color{Goldenrod}$-0.801x+0.917v+0.412$\\
		\texttt{[-0.651, 0.033]} & \color{Goldenrod}$-0.350x-0.09v+0.7741$\\
		\texttt{[-0.759, 0.018]} & \color{Goldenrod}$-0.799x+0.284v+0.395$\\
		\texttt{[-0.534, -0.020]}& \color{BlueViolet}$0.668x+0.130v-0.641$\\
		\texttt{[-0.465, 0.039]} & \color{Goldenrod}$-0.155x+0.835v+0.602$ \\						
		\hline
	\end{tabular}
	\caption{All codewords and subpolicies, rounded to 4 digits. The colors of the functions represents the ones in figure \ref{fig:mountaincar_polviz_ipr}}
\end{table}
\label{appendix:mountaincar_functions}

\section{Validation spread metrics}

\subsection{SimpleGoal}

\setlength{\tabcolsep}{12pt}
\begin{table}[H]
	\centering
	\begin{tabular}{|c|c|c|}
		\hline
		Metric      & DRL      & Ours       \\
		\hline
		Min     	& -10.0    & -11.790    \\
		Max    		& 22.667   & 22.798     \\
		Mean 		& 16.599   & 12.657     \\
		Std         & 3.117    & 8.436      \\
		Q1    		& 14.542   & 12.298     \\
		Median/Q2	& 16.956   & 15.382     \\
		Q3			& 18.687   & 17.646     \\
		IQR         & 4.145    & 5.348      \\
		Coverage	&$97.6\%$  & $86.17\%$  \\
		
		\hline
	\end{tabular}
	\caption{Spread of returns for experiments on SimpleGoal.}
\end{table}
	\label{appendix:simplegoal_spread}

\subsection{MountainCar}

\setlength{\tabcolsep}{12pt}
\begin{table}[H]
	\centering
	\begin{tabular}{|c|c|c|}
		\hline
		Metric      & DRL      & Ours       \\
		\hline
		Min     	& 93.360   & 71.050     \\
		Max    		& 93.622   & 94.680     \\
		Mean 		& 93.481   & 93.534     \\
		Std         & 0.075    & 1.345      \\
		Q1    		& 93.403   & 93.403     \\
		Median/Q2	& 93.476   & 93.476     \\
		Q3			& 93.548   & 93.548     \\
		IQR         & 0.145    & 0.145      \\
		Coverage	&$100\%$   & $92.18\%$  \\
		
		\hline
	\end{tabular}
	\caption{Spread of returns for experiments on MountainCar.}
\end{table}
\label{appendix:mountaincar_spread}

\end{document}